\documentclass[letterpaper,10pt,conference]{IEEEtran} 

\usepackage{algorithmic}
\usepackage{amsmath,amssymb,amsfonts}
\usepackage{booktabs}
\usepackage{cite}
\usepackage[outline]{contour} 
\usepackage{graphicx}
\usepackage{lipsum}
\usepackage{tikz}
\usepackage{tikz-3dplot}
\usepackage{todonotes}
\usepackage{xcolor}
\usepackage{multirow}
\usepackage{gensymb}

\usepackage{xurl}

\usepackage[T1]{fontenc}

\usepackage[absolute,overlay]{textpos}
\usepackage{tikz}

\usepackage[acronym]{glossaries}
\newacronym{GPS}{GPS}{global positioning system}
\newacronym{GNSS}{GNSS}{global navigation satellite system}
\newacronym{RTK}{RTK}{real-time kinematic}
\newacronym{IR}{IR}{infrared}
\newacronym{DoF}{DoF}{degrees of freedom}
\newacronym{IMU}{IMU}{inertial measurement unit}
\newacronym{AGL}{AGL}{altitude above ground level}
\newacronym{LiDAR}{LiDAR}{light detection and ranging}

\colorlet{veccol}{green!50!black}
\colorlet{projcol}{blue!70!black}
\colorlet{myblue}{blue!80!black}
\colorlet{myred}{red!90!black}
\colorlet{mydarkblue}{blue!50!black}
\tikzset{>=latex} 
\tikzstyle{proj}=[projcol!80,line width=0.08] 
\tikzstyle{area}=[draw=veccol,fill=veccol!80,fill opacity=0.6]
\tikzstyle{vector}=[-stealth,myblue,thick,line cap=round]
\tikzstyle{unit vector}=[->,veccol,thick,line cap=round]
\tikzstyle{dark unit vector}=[unit vector,veccol!70!black]
\usetikzlibrary{angles,quotes} 
\contourlength{1.3pt}
\def\BibTeX{{\rm B\kern-.05em{\sc i\kern-.025em b}\kern-.08em
    T\kern-.1667em\lower.7ex\hbox{E}\kern-.125emX}}
\begin{document}

\title{\huge
	A Precision Drone Landing System using Visual and IR Fiducial Markers and a Multi-Payload Camera
}

\author{\IEEEauthorblockN{Joshua Springer}
\IEEEauthorblockA{\textit{Department of Computer Science} \\
\textit{Reykjavik University}\\
Reykjavik, Iceland \\
joshua19@ru.is}
\and
\IEEEauthorblockN{Gylfi Þór Guðmundsson}
\IEEEauthorblockA{\textit{Department of Computer Science} \\
\textit{Reykjavik University}\\
Reykjavik, Iceland \\
gylfig@ru.is}
\and
\IEEEauthorblockN{Marcel Kyas}
\IEEEauthorblockA{\textit{Department of Computer Science} \\
\textit{Reykjavik University}\\
Reykjavik, Iceland \\
marcel@ru.is}
}

\maketitle


\begin{abstract}
	We propose a method for autonomous precision drone landing with fiducial markers
and a gimbal-mounted, multi-payload camera with wide-angle, zoom, and IR sensors.
The method has minimal data requirements;
it depends primarily on the direction from the drone to the landing pad,
enabling it to switch dynamically between the camera's different sensors and zoom factors,
and minimizing auxiliary sensor requirements.
It eliminates the need for data such as altitude above ground level,
straight-line distance to the landing pad, fiducial marker size,
and 6 DoF marker pose (of which the orientation is problematic).
We leverage the zoom and wide-angle cameras, as well as
visual April Tag fiducial markers to conduct successful precision landings
from much longer distances than in previous work (168m horizontal distance, 102m altitude).
We use two types of April Tags in the IR spectrum -- active and passive --
for precision landing both at daytime and nighttime,
instead of simple IR beacons used in most previous work.
The active IR landing pad is heated;
the novel, passive one is unpowered, at ambient temperature,
and depends on its high reflectivity and an IR differential between the ground and the sky.
Finally, we propose a high-level control policy to manage initial search for the landing pad
and subsequent searches if it is lost -- not addressed in previous work.
The method demonstrates successful landings with the landing skids at least touching the landing pad,
achieving an average error of 0.19m.
It also demonstrates successful recovery and landing when the landing pad is temporarily obscured.


\end{abstract}


\section{Introduction}
Takeoff and waypoint-to-waypoint navigation have been reliably automated in drone flight,
with \gls{GPS} as the main navigational tool.
Drones can generally execute missions safely
even with typical \gls{GPS} errors of 1m or more~\cite{gps_error}
because they tend to operate in free airspace away from obstacles.
However, this error becomes problematic during the landing process,
when the drone may have to operate in a constrained space due to obstacles or other restrictions.
While it is possible to decrease \gls{GPS} error using \gls{RTK} systems,
these are not available at every landing site,
and \gls{GPS} itself is not even always available.
Furthermore, \gls{GPS}-based landings are blind to the environment
and therefore cannot intelligently avoid obstacles.

A common way of improving \gls{GPS} landings is to mark the landing pad with an indicator
that makes it possible to detect with sensors onboard the drone,
and then carry out a landing precisely on the landing pad.
One such indicator is a \emph{fiducial marker}
-- a 2D, high-contrast pattern that marks significant objects or locations.
Fiducial systems such as 
April Tag~\cite{apriltag3}, AR Tag~\cite{artag}, ArUco~\cite{aruco}, and WhyCode~\cite{whycode}
provide
reliable, low false-positive identification of distinct markers via an encoded ID number.
They extract the \emph{pose} (position and orientation) of the markers relative to a camera
using monocular images, the camera's distortion parameters,
and the size of the fiducial marker.
While the position is straightforward to compute,
the orientation is subject to ambiguity~\cite{pose_ambiguity},
leading to erroneous output~\cite{irc_evaluation_orientation_ambiguity}.
Similarly, \gls{IR} beacons actively create a signal that can be recognized using an \gls{IR}
camera, which is also a common drone peripheral sensor.
\Gls{IR} beacons provide a means of estimating the pixel position of a landing pad in a camera feed,
and in some cases provide a distance estimate.

A simple way of identifying landing pads marked with fiducial markers or \gls{IR} beacons is to
mount a downward-facing camera on the drone
to determine its relative position to the landing pad
if the drone is \emph{already} in a limited space above the landing pad.
However, detecting marked landing pads is difficult if the camera is rigidly mounted to the drone
because of the drone's motion.
Further, even if it is mounted on a stabilized gimbal, the system may lose sight of the landing pad,
e.g., if the wind blows it off course.
This difficulty has led to a strategy of
actively tracking the marker independently of the drone's motion,
which increases the reliability of the detection.
However, this strategy complicates the estimation of the relative pose from the drone to the landing pad,
such that it requires more data to determine
(e.g.,
the camera's orientation,
the drone's \gls{AGL},
or
the straight-line distance from the camera to the landing pad)
or sophisticated ground infrastructure
(e.g., collaborative, intelligent landing pads or augmented fiducial systems).
Such extra requirements restrict the drone's hardware and landing behavior.

We contribute a precision drone landing method with a gimbal-mounted, multi-payload camera
that minimizes data requirements, primarily using the direction
from the drone to the landing pad as measured by the gimbal and its camera.
Previous work tends to require at least one of the following:
the drone's \gls{AGL},
the straight-line distance to the landing pad,
physical marker size, or marker pose.
Minimizing the required data enables the method
to switch between different cameras
and landing pads with minimal reconfiguration,
whereas previous work depends on static systems.
We use a zoom camera for long-range detection,
a wide-angle camera for short-range detection,
and an \gls{IR} camera for both daytime and nighttime detection.
We mark landing pads with fiducial markers (April Tags) in both the visual and \gls{IR} spectra
(instead of previously-used \gls{IR} \emph{beacons}),
allowing the system to distinguish between different landing pads in the \gls{IR} spectrum reliably.
We test an active, heated IR April Tag
and a novel, unheated, high-reflectivity April Tag for landing infrastructure that is usable at nighttime
without power.
Finally, we develop a control policy for approaching and landing on the landing pad
that manages the initial search for the landing pad
and subsequent searches if the landing
pad is lost during approach or descent.


\section{Related Work}
\label{section:related_work}

The most basic paradigm of autonomous landing with a camera
and a marked landing pad assumes the camera has a downward-facing orientation,
either rigidly mounted to the drone or stabilized vertically down on a gimbal.
Araar
et al.~use downward-facing cameras and the 6~\gls{DoF} pose of an April Tag,
encountering issues of orientation ambiguity, and solving them by using many markers and a voting
scheme to decide the position of the landing pad~\cite{vision_based_moving}.
Olivarez-Mendez et al.~use a camera fixed to a helicopter to locate a custom visual marker
on a stationary platform, approaching from a distance of 4m
and achieving an error of 0.7m~\cite{helicopter_fuzzy_landing}.
Badakis et al.~use a downward facing camera to locate 
an ArUco marker with an embedded \gls{IR} beacon
and measure the drone's height using a barometer and rangefinder.
They achieve an average error of 0.15m when using both the visual and \gls{IR} systems,
and an error of 0.25m when using only one system~\cite{landing_downward_camera_fractal_ir}.
Nowak et al.~use an \gls{IR} beacon on the landing pad
and \gls{LiDAR} to determine the drone's height, achieving successful landing on a 0.4m by 0.4m
platform~\cite{landing_downward_camera_ir_beacon_groundbased}.
Xuan-Mung et al.~use an \gls{IR} beacon mounted
on a landing pad that is heaving to mimick a ship's motion
and
detect it using a downward-facing \gls{IR}
camera~\cite{landing_downward_camera_ir_beacon_lidar_height_heaving_deck}.
Instead of using a marker or beacon,
Pluckter and Scherer record the appearance of the takeoff location
and realign the drone with it using visual matching
after executing a mission~\cite{scherer_landing_visual_teach_repeat_fisheye}.
Polvara et al.~use a fixed, forward-facing camera for initial approach,
and a fixed, downward-facing camera for final descent.
Their system collaborates with the moving landing pad (marked with an AR Tag)
to share \gls{GPS} position, velocity, and acceleration data
(in simulation)~\cite{multirotor_landing_moving_vessel}.
Falanga et al.~use a custom marker, one downward-facing camera, and one forward-facing
camera to land on a 1.5m by 1.5m platform marked
with a custom visual marker~\cite{vision_based_moving_2}.
While this paradigm is employed successfully, it suffers from the basic weakness that it is difficult
to detect and maintain detection of the marker if the camera is fixed or only pointing downward
because the camera moves according to the movements of the drone.
Additionally, the camera's limited field of view makes it difficult to find the marker if it
cannot move independently.

Another paradigm of autonomous landing uses a gimbal-mounted camera that is actuated to track
the landing pad independently of the drone's orientation,
which increases the robustness of the landing pad detection
but complicates the landing process.
Borowczyk et al.~present a hybrid method
that uses a collaborative \gls{GPS} system (both on the moving landing pad and the drone)
for long-range approach,
a gimbal-mounted camera for initial visual approach,
and a downward-facing, wide-angle camera for final
descent~\cite{multirotor_landing_high_velocity_ground_vehicle}.
Cho et al.~use a gimbal-mounted camera to detect the 6~\gls{DoF} pose of an
AR Tag on a moving ship deck, initially approaching using the \gls{GPS} data of both the ship
and the drone, and achieving an error of 0.2m~\cite{landing_uav_ship_deck_servoing}.
Jiang et al.~use a gimbal-mounted camera to track a landing pad with a custom visual marker
and approach it using the gimbal angles and height as measured by an altitude
sensor~\cite{landing_gimbal_camera_custom_visual_marker_distance_height}.
Demirhan and Premachandra use a gimbal-mounted camera
to detect a custom visual marker and approach using
the camera's orientation,
the marker's pose, and the drone's \gls{AGL},
measured by an auxiliary sensor~\cite{landing_servo_gimbal_micro_uav_custom_marker_height_sensor}.
Lim et al.~use an omnidirectional \gls{IR} beacon
mounted on the landing pad
and approach it using the gimbal-mounted camera's attitude
and the drone's height as measured from a
\gls{LiDAR}~\cite{landing_gimbal_ir_omnidirectional_lidar_height}.
They can detect and approach from a distance of 18m.
Tanaka et al.~use the full 6 \gls{DoF} pose of the marker -- normally subject to orientation ambiguity --
by creating a proprietary version of AR Tag called Lentimark~\cite{lentimark},
which has wave-like patterns on its side to make the rotation unambiguous~\cite{lentimark_landing}.
Our previous studies evaluate several fiducial systems
to empirically test the effect of pose ambiguity
on the behavior of the drone~\cite{irc_evaluation_orientation_ambiguity,irc_autonomous_landing},
determining that it
leads to
erroneous control signals and erratic
behavior,
but that some systems still produce succesful landings.

Some studies have created fiducial markers in the \gls{IR} spectrum.
Dogan et al.~have used 3D-printed ArUco markers and QR codes to unobtrusively mark objects for augmented reality~\cite{infraredtags}.
Khattack et al.~have used laser-cut, acrylic ArUco markers for drone localization in visually
degraded environments~\cite{ir_tag_localization}.
Claro et al.~created a multimodal, active thermal marker based on ArTag for autonomous landing of a drone with \gls{GNSS}, \gls{RTK}, \gls{LiDAR} and visual and thermal cameras,
achieving an error of 0.03m and using a downward-facing camera on a stabilized gimbal~\cite{landing_ir_fiducial_marker_artuga}.
The marker is composed of two materials of differing reflectivities, with one heated and one unheated.

Most of the related work depends on the \gls{AGL} or range (straight-line distance to the landing pad)
which must be measured by auxiliary sensors, e.g., \gls{LiDAR} or ultrasonic sensors.
\Gls{AGL} may not correspond directly to the altitude above the landing pad if the terrain
is not flat, such that the system may miscalculate the position of the drone relative to the landing pad.
The method described in~\cite{landing_gimbal_camera_custom_visual_marker_distance_height}
is similar to what we propose, but we improve upon it by not requiring the height,
by adapting it to multiple cameras,
and
by using \gls{IR} fiducial markers.
Additionally
the \gls{IR} beacons are not necessarily distinguishable from each other
-- often characterized simply by areas of significant \gls{IR} radiation,
such that any \gls{IR} emission might appear as a landing pad~\cite{landing_downward_camera_ir_beacon_groundbased,landing_downward_camera_ir_beacon_lidar_height_heaving_deck}.
This issue is partially addressed in~\cite{landing_ir_fiducial_marker_artuga},
with an actively heated marker and a downward-facing camera,
but we believe there is also potential
in using actuated cameras
and
passive \gls{IR} infrastructure that exploits high reflectivity.
Finally, most of the related work does not focus on the initial search for the landing pad
or on detection re-acquisition if it is lost.
Instead, it treats autonomous landing as a low-level control problem,
detailing methods for sensor fusion and filtering to develop reliable pose estimation
and trajectory calculation.
However, many modern drones make this unnecessary by providing
\emph{position} or \emph{velocity} modes
that execute high-level commands, e.g., \emph{forward x m/s},
using either \gls{GPS} or optical flow for velocity measurement.
Furthermore, many of the methods depend on 
expensive, sophisticated, or collaborative ground infrastructure,
whereas we minimize the required data and sensors.

\section{System Overview}
\subsection{Required Data and Processing}

\begin{figure}
\centering

	\tdplotsetmaincoords{60}{50}
	\begin{tikzpicture}[scale=4,tdplot_main_coords]

		\def\rvec{1}
		\def\thetavec{45}
		\def\phivec{50}

		\coordinate (O) at (0,0,0);
		\coordinate (pxy2) at (1,0.52532198881,0);
		\draw[thin,->] (0,0,0) -- (1,0,0) node[below left=0]{$x$};
		\draw[thin,->] (0,0,0) -- (0,1.3,0) node[right=0]{$y$};
		\draw[thin,->] (0,0,0) -- (0,0,1) node[left=0,red,midway] {$a$} node[above=-1]{$z$};

		\draw[->, dashed] (0,0,0.7071) -- (0,1,0.7071) node[right=-1]{$\vec{H}$};

		\tdplotsetcoord{P}{\rvec}{\thetavec}{\phivec}
		\draw[dashed] (Pz) -- (Pxy) node[above right=0,red,midway] {$R$} node[above=0,right=0] {};
		\draw[dashed]   (O)  -- (Pxy);

		\node[left=0] at (Pz) {D};

		\filldraw [black] (Pz) circle (0.35pt);
		\filldraw [black] (Pxy) circle (0.35pt);

		\draw[->,black]   (Pxy) -- node[midway, black, below left=0.01]{$\vec{LP}$} (pxy2);

		\tdplotdrawarc[->]{(O)}{0.4}{90}{\phivec}{anchor=west}{$\phi$}
		\tdplotsetthetaplanecoords{\phivec}
		\pic [<-,draw, "$\theta$", angle eccentricity=1.5, angle radius=0.75cm] {angle = O--Pz--Pxy};

		\tdplotdrawarc[->]{(Py)}{0.55}{90}{0}{anchor=west}{$\psi$}

	\end{tikzpicture}

	\caption{
		The angles defining the control policy for the drone (D)
		to approach the landing pad (LP).
		$\phi$ represents the pan, and $\theta$ the tilt, from the drone to the landing pad.
		$\vec{H}$ represents the heading of the drone, in the direction of the $y$-axis.
		The landing pad is at the base of the vector $\vec{LP}$, and the vector extends
		in the direction of the landing pad's yaw.
		The angle $\psi$ represents the relative yaw from the drone's heading to the
		landing pad's heading.
		The altitude $a$ and range $R$ are not used or needed for this method.}

	\label{figure:approach_angles}

\end{figure}
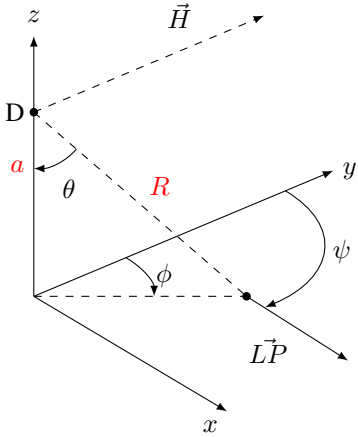

In our proposed method,
we primarily consider
the pan angle~$\phi$
and
the tilt angle~$\theta$
from the drone's heading
to the landing pad (see Fig.~\ref{figure:approach_angles}).
Secondarily,
we consider the difference $\psi$ between the drone's yaw and the landing pad's yaw,
and the pixel size $S_p$ of the landing pad (queried from the fiducial system),
and the zoom factor $Z$ (queried from the flight controller).
We use $\psi$
to point the drone
in the same direction as the landing pad
before descent,
and we direct control state transitions
and conduct intermediate calculations
based on $Z$ and $S_p$.
We avoid using the \glsreset{AGL}\gls{AGL} $a$
and the straight-line range $R$
from the drone to the landing pad because these are not available on many drones.
Additionally, the \gls{AGL} may not equal the altitude above the landing pad,
e.g., on rough terrain,
making any resulting calculations unreliable.

We determine $\phi$ and $\theta$ by tracking the landing pad with a gimbal-mounted camera
and comparing its orientation to that of the drone.
Since the gimbal is often not aimed directly at the landing pad,
but skewed by some angle as a result of the drone's motion,
we also consider the angles implied by the landing pad's pixel position in the camera frame.
Therefore, $\phi$ and $\theta$ are calculated
as in Equations~\eqref{equation:phi} and~\eqref{equation:theta},
where
$\phi_g$
and
$\theta_g$
represent the relative tilt and pan of the gimbal,
and
$\phi_u$
and
$\theta_v$
represent the additional relative pan and tilt implied
by the landing pad's pixel positions.

\begin{equation}
	\phi = \phi_g + \phi_u
	\label{equation:phi}
\end{equation}
\begin{equation}
	\theta = \theta_g + \theta_v
	\label{equation:theta}
\end{equation}

We query
$\phi_g$
and
$\theta_g$
from the flight controller
and calculate
$\phi_u$
and
$\theta_v$
by normalizing the landing pad's pixel positions
into the interval $[-0.5,0.5]$ and multiplying by the field of view in the relevant dimension,
as in Equation~\eqref{equation:phi_u}. 
Here,
$u$ is the horizontal pixel center of the landing pad,
$u_c$ is the center pixel of the camera's horizontal dimension,
and
$\text{FOV}_u$ is the camera's horizontal field of view.
We conduct a similar calculation for the vertical dimension,
with
$v$, 
$v_c$, 
and
$\text{FOV}_v$.
\begin{equation}
	\phi_u = \dfrac{u-u_c}{2u_c} \cdot \text{FOV}_u
	\label{equation:phi_u}
\end{equation}

We calculate the horizontal field of view $\text{FOV}_u$ for each camera
using Equation~\eqref{equation:fov}~\cite{physics_of_digital_photography},
where $f$ is the camera's focal length and
$w$ is its sensor width.
For cameras with a fixed focal length $F$,
we use $f=F$;
otherwise, we use $f=ZF_b$, where
$Z$ is the zoom factor
and
$F_b$ is the base focal length.
We conduct a similar calculation for the vertical field of view $\text{FOV}_v$
with the sensor height $h$ instead of $w$.

\begin{equation}
	\text{FOV}_u = 2 \arctan\dfrac{w}{2f}
	\label{equation:fov}
\end{equation}

Calculating $\phi$ and $\theta$
in this way allows the system to switch between cameras and zoom factors,
such that it can identify and track the landing pad from long distances using the zoom camera,
and then continually zoom out as it approaches.

\subsection{Control Policy}

\begin{figure}
	\centering
	\includegraphics[width=\columnwidth]{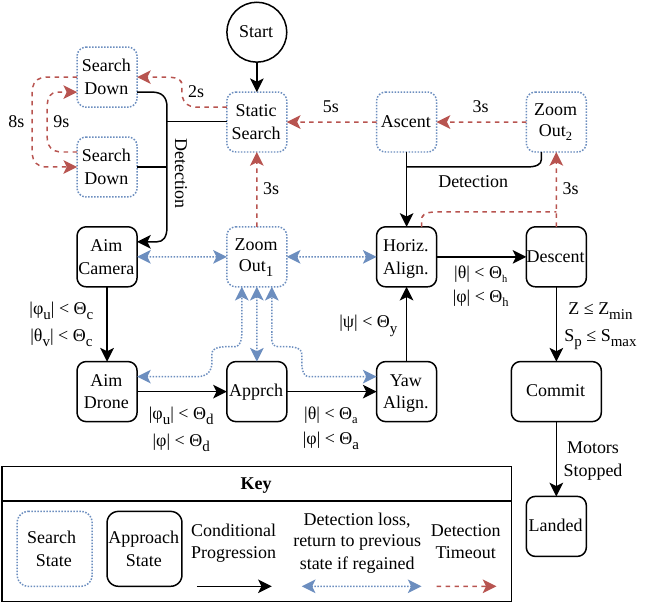}
	\caption{
		The control policy.
		For \emph{Aim Camera}, \emph{Aim Drone}, and \emph{Approach},
		transition conditions are based on
		$\Theta_c = \Theta_d = \Theta_a = 3\degree$.
		For \emph{Yaw Alignment}, $\Theta_y = 1\degree$.
		For \emph{Horizontal Alignment}, $\Theta_h = 2\degree$.
		For \emph{Descent}, $Z_\text{min}=2$, and $S_\text{max}=32\%$.
		For \emph{Commit}, the flight controller indicates motor stop.
	}

	\label{figure:control_policy}
\end{figure}

The autonomous landing system follows the control policy outlined in
Fig.~\ref{figure:control_policy}.
We assume the drone starts the landing process near
-- but not directly above -- the landing pad,
having navigated via \gls{GPS} or some other means.
It starts to look for the marker by holding the camera still,
facing forward and down (\emph{static search}).
It then proceeds to search in a slow, clockwise yaw rotation,
tilting the camera down (\emph{search down}) and up (\emph{search up}).
If the system detects the landing pad in any of the search states,
it centers the landing pad in the camera's field of view (\emph{aim camera}),
and then points the drone at the landing pad in the yaw dimension (\emph{aim drone}).
At this point, the drone can \emph{approach} simply by moving forward,
making slight left or right adjustments, and tracking the landing pad with the camera.
Once the drone is above the landing pad,
it stops actively tracking the marker
(but continues to zoom in or out to keep the landing pad in the camera frame),
points the gimbal vertically down,
and
aligns to the landing pad's yaw (\emph{yaw align}).
It then aligns horizontally
(\emph{horizontal alignment})
and begins its \emph{descent}.
Once the zoom factor is adequately low and the pixel size of the landing pad is adequately high,
the drone \emph{commits} to the landing,
descending until hitting the ground and disabling the motors.
It then transitions to the \emph{landed} phase.
If the landing pad is no longer detected at any point during
\emph{aim camera},
\emph{aim drone},
\emph{approach},
\emph{yaw alignment},
or
\emph{horizontal alignment},
the system zooms out (\emph{zoom out$_1$}),
to reacquire the landing pad.
If it succeeds, it switches back to the previous state and resumes the landing process;
otherwise, it restarts from \emph{static search}.
If the landing pad is no longer detected during the \emph{descent} stage,
the drone recovers by zooming out (\emph{zoom out$_2$}) and, if necessary, ascending (\emph{ascent}),
and returning to \emph{horizontal alignment} if it reacquires it;
otherwise, it restarts from \emph{static search}.

\begin{table*}[t]
	\centering
	\caption{Control Signal Calculation}
	\label{table:control_signal_calculation}
	\begin{tabular}{p{2cm}p{2cm}p{2cm}p{2cm}p{2cm}p{1cm}p{1cm}p{1cm}}
\toprule
                Mode &       Forward (m/s) &       Right (m/s) & Up (m/s) & Yaw (\degree/s, CW) &                    Pan &                    Tilt & Zoom \\
\midrule
       Static Search &                   0 &                 0 &      0.0 &                   0 &          $0~\degree/s$ &           $0~\degree/s$ & none \\
         Search Down &                   0 &                 0 &      0.0 &                   5 &          $0~\degree/s$ &         $-10~\degree/s$ & none \\
           Search Up &                   0 &                 0 &      0.0 &                   5 &          $0~\degree/s$ &          $10~\degree/s$ & none \\
          Aim Camera &                   0 &                 0 &      0.0 &                   0 & $1.2 \theta_u~\degree$ & $-1.2 \theta_v~\degree$ & auto \\
           Aim Drone &                   0 &                 0 &      0.0 &              $\phi$ & $5 \theta_u~\degree/s$ &  $5 \theta_v~\degree/s$ & auto \\
            Approach & $2.82 \sin(\theta)$ & $1.41 \sin(\phi)$ &      0.0 &                   0 & $5 \theta_u~\degree/s$ &  $5 \theta_v~\degree/s$ & auto \\
       Yaw Alignment &                   0 &                 0 &      0.0 &             $-\psi$ &             $0\degree$ &              $0\degree$ & auto \\
Horizontal Alignment &    $-0.10 \theta_v$ &   $0.10 \theta_u$ &      0.0 &                   0 &             $0\degree$ &              $0\degree$ & auto \\
             Descent &    $-0.05 \theta_v$ &   $0.05 \theta_u$ &     -0.5 &                   0 &             $0\degree$ &              $0\degree$ & auto \\
              Commit &                   0 &                 0 &     -0.5 &                   0 &             $0\degree$ &              $0\degree$ & none \\
              Landed &                     &                   &          &                     &                        &                         &      \\
            Zoom Out &                   0 &                 0 &      0.0 &                   0 &          $0~\degree/s$ &           $0~\degree/s$ &  out \\
              Ascent &                   0 &                 0 &      0.5 &                   0 &          $0~\degree/s$ &           $0~\degree/s$ & none \\
\bottomrule
\end{tabular}

\end{table*}

Each state of the control policy has parameters to specify
velocity targets for the drone in the forward, up, and right dimensions,
clockwise yaw rate,
and
the target pan and tilt speed (or angle) of rotation of the camera,
as well as the mode's zoom policy
(see Table~\ref{table:control_signal_calculation}).
The pan and tilt are controlled in terms of rotational speed in modes where the landing pad
is being actively tracked,
and in terms of angles when the camera should be pointed straight ahead, straight down,
or upon initial recognition of the marker
(when it is advantageous to immediately aim the camera directly at the marker without constraining
pixel speed).
The control signals are constrained before execution as follows:
the forward command is in $[-0.5, 2.0]$ m/s,
the right command is in $[-1.0, -1.0]$ m/s,
the up command is in $[-0.5, 1.0]$ m/s,
and
the yaw rate command is in $[-10, 10]$ \degree/s.
For the zoom modes,
the mode \emph{none} means the zoom factor does not change,
the mode \emph{out} means the zoom factor decreases slowly,
and
the mode \emph{auto} means the system minimally adjusts the zoom factor to keep
the pixel size of the landing pad between 20\% and 80\% of the total video stream pixel size.
Pixel sizes out of this range make the detection unreliable:
if the pixel size is too small, the system may not detect it;
if it is too large, the drone's movements can push the marker out of the
camera's field of view.
We minimally adjust the zoom factor because zooming increases pixel speed and blur,
making the detection less reliable.
Changing the zoom factor over the lower-bound of the zoom camera triggers
a stream change to the wide-angle camera or vice versa.
Zoom policies do not apply when using the \gls{IR} camera, as it has no optical zoom.

\subsection{Experimental Infrastructure}

\begin{figure}
        \centering
        \includegraphics[height=6cm]{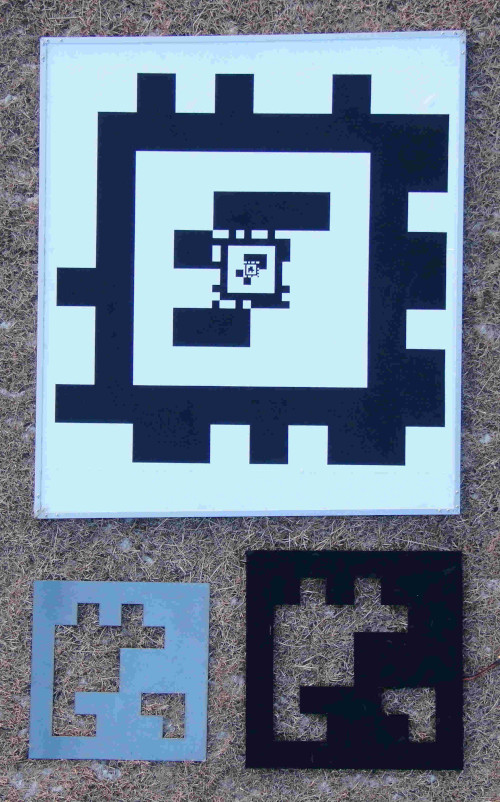}
        \includegraphics[height=6cm]{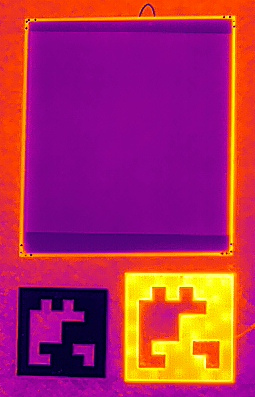}
	\caption{RGB and \gls{IR} pictures of the landing pads, taken from the drone.
	Top: visual landing pad.
	Bottom left: passive \gls{IR} landing pad (at ambient temperature, high reflectivity).
	Bottom right: active (heated) \gls{IR} landing pad.
	}
        \label{figure:all_landing_pads_6_december_2023}
\end{figure}

\begin{figure}
	\centering
	\includegraphics[width=0.7\columnwidth]{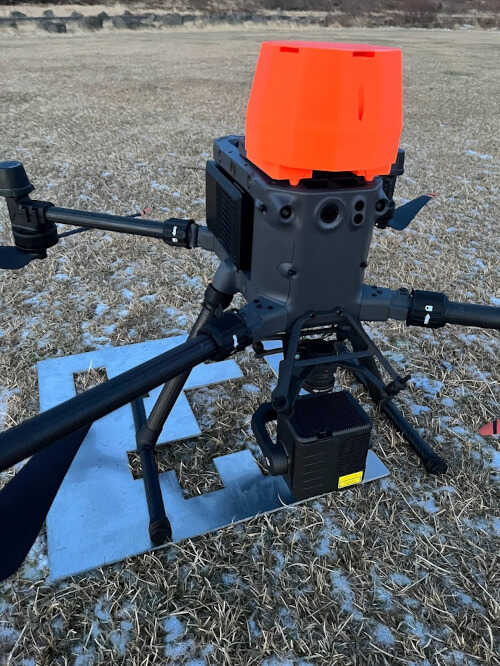}
	\caption{
		After experimental landing:
		the Matrice 350, H20T and Raspberry Pi (inside the top-mounted case)
		on top of the passive \gls{IR} landing pad.
		The landing pad is placed on a backdrop of grass with small patches of snow.
	}
	\label{figure:experimental_infrastructure}
\end{figure}

We conduct real-world landing experiments to test the accuracy of the method.
We use 3 different landing pads,
as shown in Fig.~\ref{figure:all_landing_pads_6_december_2023}.
To test the method's basic and long-range performance,
we use a visual landing pad
with 3 concentric April Tag markers.
The smaller markers allow the system to detect the landing pad at low altitudes.
We test two \gls{IR} landing pads for nighttime landing,
both in the shape of an April Tag 36h11 with ID 564
because its lack of islands and thin corners allow it to be reliably cut out of a single,
contiguous piece of metal.
The actively-heated \gls{IR} landing pad (right)
is made out of steel with vinyl tape on the top
to
increase its emissivity.
It emits relatively high \gls{IR} radiation compared to the ground
because of its actual temperature difference~\cite{overview_of_radiation_thermometry}.
Since it emits \gls{IR} radiation in the dark regions
of the corresponding April Tag marker,
we invert the IR stream before passing it to the April Tag detector.
Heating the landing pad is an intuitive way of increasing its \gls{IR} radiation
but requires a power source.
For this reason, we also test a passive aluminum \gls{IR} landing pad (left),
which is at ambient temperature when in use and requires no power source.
Its high reflectivity 
means that it reflects the \gls{IR} radiation from the sky,
which is relatively low compared to the ground,
and thus the landing pad appears cold and dark.
The differences in the sizes of the landing pads are due only to incidental availability of materials.
We test on a DJI Matrice 350~\cite{matrice_350_rtk_user_manual_2023}
with an H20T~\cite{zenmuse_h20_series_user_manual_2020},
which has wide-angle, zoom, and \gls{IR} cameras.
We implement the control policy
on a custom computational payload
(see Fig.~\ref{figure:experimental_infrastructure})
containing a Raspberry~Pi~4
that is mounted onboard the drone
and runs a DJI Payload SDK application to handle
video decoding,
data subscription,
camera management,
gimbal control,
flight control,
and widgets for interaction 
with 
the pilot~\cite{landing_psdk_app_repo}.

\section{Results}
\label{section:results}

We conducted 26 tests over 4 days with air temperatures ranging from $-8^\circ$C to $3^\circ$C,
wind conditions less than 5 m/s,
on a grassy field in Iceland with light snow.
We started each test by manually positioning the drone
at a horizontal distance between 5m and 168m from the landing pad,
and altitude of 5m to 102m above the landing pad.
We then gave control of the drone to the computational payload,
which directed the landing according to the policy in Fig.~\ref{figure:control_policy},
while we simply monitored its behavior.
The error of the landing is measured as the distance from the center of the
landing pad to the point directly under the drone's camera after touchdown.
Sample landing videos can be found on Vimeo~\cite{vimeo_matrice_fiducial_landings}.

\begin{table}[]
	\caption{Distribution of Landing Errors}
	\centering
	\resizebox{\columnwidth}{!}
	{
		\begin{tabular}{lrrrrr}
\toprule
      Type &  $\mu_E$ (m) &  $\sigma_E$ (m) &  $n$ &  Max Alt (m) &  Max Dist (m) \\
\midrule
    Visual &         0.16 &            0.10 &   11 &          102 &           168 \\
 Active IR &         0.14 &            0.14 &    5 &           15 &            15 \\
Passive IR &         0.26 &            0.17 &   10 &           15 &            14 \\
       All &         0.19 &            0.14 &   26 &          102 &           168 \\
\bottomrule
\end{tabular}

	}
	\label{table:error_meta}
\end{table}

Table~\ref{table:error_meta} shows the distribution of the landing error, where
$\mu_E$ is the average error,
$\sigma_E$ is the standard deviation of the error,
$n$ is the number of tests per type, and
Max Alt. and Max Dist. are the maximum starting altitudes and horizontal distances
for the landing tests.
All tests were successful, i.e., the drone landed at least touching the landing pad,
with an average error of 0.19m,
which is lower than most of the previous work
-- except those using sophisticated systems
e.g., \gls{RTK} and \gls{LiDAR} --
and does not depend on the commonly-needed \gls{AGL} $a$ or range $R$ from Fig.~\ref{figure:approach_angles}.
This demonstrates the viability of the method and each of the landing pad types.

\begin{figure}[]
        \centering
        \includegraphics[height=2.5cm]{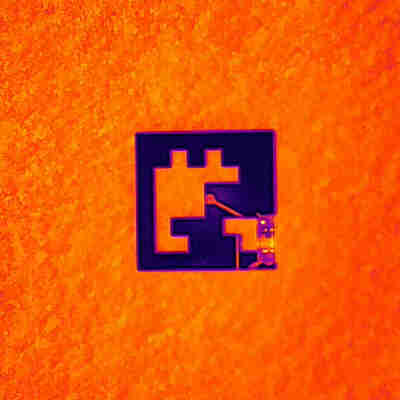}
        \includegraphics[height=2.5cm]{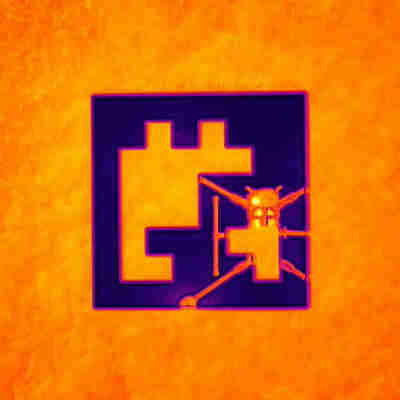}
        \includegraphics[height=2.5cm]{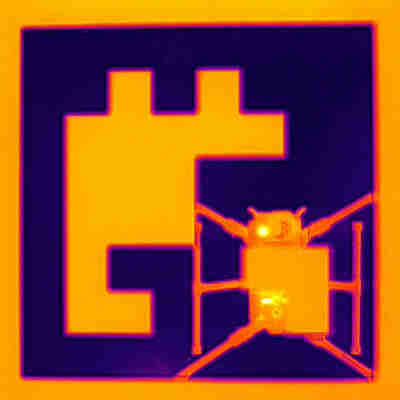}
        \caption
	{
		Reflection of the drone's heat signature in the passive \gls{IR} landing pad.
		Left: 13m altitude. Middle: 10m altitude. Right: 7m altitude.
	}
        \label{figure:passive_thermal_landing_pad_progression}
\end{figure}

The system carries out landings from much longer distances than previously reported
using the visual landing pad and zoom camera.
The \gls{IR} camera's range is limited by its lack of zoom,
but these experiments demonstrate the viability of using
both active and passive \gls{IR} fiducial markers for autonomous
landing.
The active \gls{IR} landing pad
provides consistent performance and is clearly recognizable.
The passive \gls{IR} landing pad requires no power source but reflects the drone's IR radiation
(see Fig.~\ref{figure:passive_thermal_landing_pad_progression}),
causing occlusion when horizontally aligned at low altitudes and
necessitating the \emph{commit} state.
The performance of the passive \gls{IR} landing pad may be weather-dependent
and requires further testing.
Since the the control policy is the same for all landing pads,
the only major performance difference is the maximum detection distance.

\subsection{Case Study: Antagonistic Landing Site}

\begin{figure}[h]
        \centering
	\includegraphics[width=0.9\columnwidth]{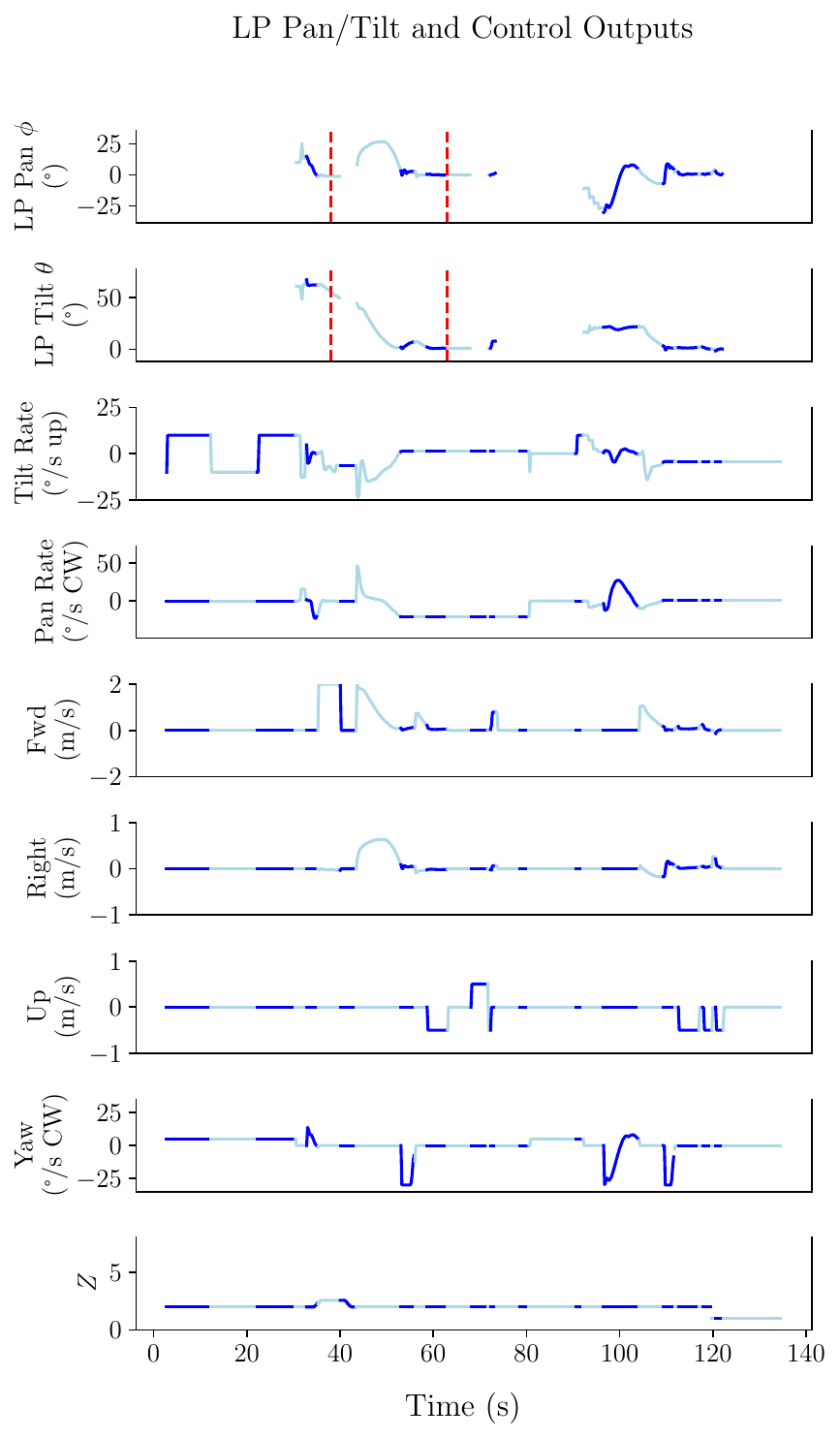}
	\caption
	{
		To illustrate the control policy's robustness against detection losses:
		a landing with the visual landing pad and zoom camera,
		with simulated obstacles and wind gusts by via intentional obscuration
		of the landing pad.
		Color changes indicate mode changes;
		vertical lines indicate intentional obscuration.
	}
        \label{figure:example_landing_with_corrections}
\end{figure}

Successes in the first experiment led to a more antagonistic second experiment where the landing pad
was intentionally obscured both during the approach and descent states
to test the system's ability to recover from detection loss.
Fig.~\ref{figure:example_landing_with_corrections}
illustrates this,
showing the control signals for a landing with the
visual landing pad and the zoom camera.
The system reacquires the landing pad each time by zooming out or ascending to increase its field of
view,
restarts the search process entirely when necessary,
and
ultimately lands successfully.
The drone starts at a distance of 20m and altitude of 10m relative to the landing pad.
It searches for the landing pad by tilting the gimbal up and down while yawing clockwise
from the start until $t=30$.
Then, it detects the landing pad, 
aiming the camera at it until $t=32$
and aiming the drone at it until $t=35$.
It begins its approach until the landing pad is intentionally obscured and thrown to the
side at $t=40$, at which point it zooms out and finds the landing pad towards its right.
It begins its approach again at $t=44$,
until arriving above it at $t=53$,
when it turns counterclockwise to align with the landing pad's yaw.
It conducts a horizontal alignment at $t=56$ and begins its descent at $t=58$.
At $t=63$, the landing pad is again intentionally obscured and thrown to the side,
causing the drone to enter the \emph{zoom out 2} state
At $t=68$, it begins its ascent and momentarily re-acquires the marker at $t=71$.
It zooms out and begins the search process again at $t=78$,
first with \emph{static search} until $t=80$,
then \emph{search down} until $t=90$,
and \emph{search up} until $t=92$,
at which point it re-acquires the marker.
It aims the camera again until $t=96$,
then approaches until $t=109$,
then aligns to the landing pad's yaw until $t=111$.
It alternates between \emph{horizontal alignment} and \emph{descent}
3 times until $t=122$, when it is gradually getting closer to the landing pad
in the horizontal and vertical dimensions,
and during this time it switches to the wide-angle stream with $Z=1$.
Finally, it commits to the landing and descends vertically until it touches down
on the landing pad and disables the motors at $t=134$.
This demonstrates that the system is able to recover from detection losses by zooming out, ascending,
or restarting the search process.

\section{Conclusions \& Future Work}
We have presented a method of autonomous precision drone landing that can use
either
visual
or
\gls{IR} fiducial markers
and achieves an average error of 0.19m.
It has fewer data requirements than previous work, i.e.,
the direction towards the landing pad,
the landing pad's pixel size,
and the camera's zoom factor;
it does not use the drone's height or distance relative to the landing pad.
It avoids 6~\gls{DoF} fiducial pose estimation,
eliminating the orientation ambiguity problem from our previous study~\cite{irc_autonomous_landing}.
The method is successful from much longer distances than previous work,
with a zoom camera and visual fiducial markers.
It works at both daytime and nighttime using
actively-heated
and passive (unpowered), high-reflectivity \gls{IR} fiducial markers.
Finally,
the control policy enables the drone to search for the landing pad
and reacquire it if it is lost,
making the landings more reliable.

Further testing of this method includes
testing different
marker systems,
weather conditions,
and background surfaces.
Using a PID system (instead of a simple P system) to calculate velocity targets
would allow the method to generalize to moving landing pads.


\bibliographystyle{IEEEtran}
\bibliography{references}

\end{document}